\documentclass{article}
\usepackage{spconf,amsmath,epsfig}

\usepackage{array}
\usepackage{multirow}
\usepackage{subfigure}
\usepackage{url}
\usepackage{setspace}

\newcolumntype{V}{>{$\vcenter\bgroup\hbox\bgroup}c<{\egroup\egroup$}}
\def\Hline{\noalign{\hrule height 4\arrayrulewidth}}

\graphicspath{{./imgs/}}

\begin{document}\sloppy

\def\x{{\mathbf x}}
\def\L{{\cal L}}

\title{VIEWPOINT DISTORTION COMPENSATION IN PRACTICAL SURVEILLANCE SYSTEMS}
%
\name{Ognjen~Arandjelovi\'c, Duc-Son~Pham, and Svetha~Venkatesh}
\address{Centre for Pattern Recognition and Data Analytics, Deakin University, Australia}

\maketitle

\begin{abstract}
Our aim is to estimate the perspective-effected geometric distortion of a scene from a video feed. In contrast to all previous work we wish to achieve this using from low-level, spatio-temporally local motion features used in commercial semi-automatic surveillance systems. We: (i) describe a dense algorithm which uses motion features to estimate the perspective distortion at each image locus and then polls all such local estimates to arrive at the globally best estimate, (ii) present an alternative coarse algorithm which subdivides the image frame into blocks, and uses motion features to derive block-specific motion characteristics and constrain the relationships between these characteristics, with the perspective estimate emerging as a result of a global optimization scheme, and (iii) report the results of an evaluation using nine large sets acquired using existing close-circuit television (CCTV) cameras. Our findings demonstrate that both of the proposed methods are successful, their accuracy matching that of human labelling using complete visual data.\end{abstract}
\vspace{-5pt}\section{Introduction}\label{s:intro}\vspace{-2pt}
In this paper we address the problem of inferring the perspective of a static scene with moving objects. In its general form, this problem has been extensively studied in the past. However, unlike in previous work, here our aim is to accomplish this under a set of constraints which have not been considered previously in the literature. Specifically, we wish to estimate the dominant perspective of a scene using low-level motion features, local both in spatial and temporal sense.

Our motivation for this problem and the source of the outline constraints stems from the operational requirements of many semi-automatic CCTV-based surveillance systems~\cite{PhamAranVenk2015,AranPhamVenk2015}. In particular, we are interested in systems which occupy the largest portion of the commercial market today, and which detect abnormalities in video feeds on a semantically low-level (i.e.\ without `understanding' or interpreting the nature of the detected abnormality). They accomplish this by extracting low-level features from which a scene-specific statistical model of normal motion in a scene is learnt. Spatio-temporally local motion is universally used as the basis for the elementary features over which learning is performed. These features have been proven successful in practice and they fit the computational constraints imposed by the need for most of the processing and data storage to be done locally using the camera's on-board hardware. This particular constraint emerges as a consequence of the large scale of many practical CCTV systems which for scalability reasons have a distributed architecture with minimal reliance on communication with the central hub.

As detailed in the next section, none of the previous work has addressed the problem of perspective estimation using the setup described above. In this paper we introduce two algorithms that solve the problem using two different approaches.

\vspace{-5pt}\section{Related work}\label{s:prev}\vspace{-2pt}
The problem addressed in the present paper is intimately related with the corpus of work on estimating the position and orientation of a camera and the recovery of 3D scene structure. This has been an active topic of research since the early days of computer vision, resulting in a number of mature techniques which are now widely used in practice (in film production, for example).

When there are known correspondences between 3D world points and their 2D projections, there is a series of algorithms that have been described in the literature which successfully handle cases for different numbers of available correspondences using different (usually iterative) optimization schemes \cite{KahlHenr2007,HescRoum2011}. When no explicit world-to-camera mapping data is available but the camera is in motion and the scene mostly static, perceived (image) motion and different types of constraints (e.g.\ probabilistic, epipolar, or motion parallax based) can be used instead \cite{ArmaArajSalv2003,DomkAloi2007}. Stronger yet constraints must be employed when it is not possible to obtain 3D-to-2D correspondences and the camera is static. For example for built-up scenes outline maps of buildings have been used with success by a number of researchers \cite{ChamCiptTanPham+2010}. Similarly, in urban or indoor scenes the presence of many parallel lines (e.g\ corridor or street boundaries) and their convergence towards the same vanishing point can be used to estimate the perspective \cite{BaziSeoDemoVass+2012,HeroSzenZachDubs+2013}. Yet others learn appearance of different types of elementary structures which allows them to build an approximate model of the scene \cite{HoieEfroHebe2005,CornLeibCornVanG2008}.

As we shall see in the next section, none of these approaches meet all the requirements of a practical CCTV system. The cameras we are dealing with are static, their large number makes an elaborate setup procedure required to obtain 3D-to-2D point correspondences impractical, and the algorithm deployed must be sufficiently robust to handle a wide variety of scenes and poor image/video quality thus prohibiting the reliance on the presence of strong cues such as parallel lines or specific object types.

\vspace{-5pt}\section{Background -- operational constraints}\label{s:constraints}\vspace{-2pt}
Computer-assisted video surveillance data analysis is of major commercial and law enforcement interest. Unsurprisingly, this interest has spurred a major research effort as well; topics such as human detection~\cite{DalaTrig2005} and tracking~\cite{MartAran2010}, crowd analysis~\cite{Aran2008}, activity recognition~\cite{TranSoro2008}, and others, have been attracting a significant amount of attention. Nonetheless, commercially available systems are still in relative infancy with most of the methods described in the academic literature still not sufficiently mature for practical deployment. Achieving reliable performance for scenes of different types, a wide range of foreground object classes which themselves exhibit major within-class variability, changing illumination conditions and shadows, and the breadth of possible events of interest all continue to pose a major challenge.

On a broad scale, systems currently available on the market can be grouped into two categories in terms of their approach. The first group focuses on a relatively small, predefined and well understood subset of events or behaviours of interest \cite{Phil,LaveKhanThur2007}. Examples include the detection of unattended baggage, violent behaviour, or specific incidences of vandalism. While suitable in certain circumstances, the narrow focus of these systems prohibits their applicability in less constrained environments in which a more general capability is required. In addition, these approaches tend to be computationally expensive and error prone, often requiring fine tuning by skilled technicians. This is not practical in many circumstances, for example when hundreds of cameras need to be deployed as is often the case with CCTV systems operated by municipal authorities. The second group of systems approaches the problem of detecting suspicious events at a semantically lower level \cite{Aran2011a,Inte,AranPhamVenk2015d,iCet}. Their central paradigm is that an unusual behaviour at a high semantic level will be associated with statistically unusual patterns (also `behaviour' in a sense) at a low semantic level -- the level of elementary image/video features. Thus methods of this group detect events of interest by learning the scope of normal variability of low-level patterns and alerting to anything that does not conform to this model of what is expected in a scene, without `understanding' or interpreting the nature of the event itself. These methods nearly universally employ motion features~\cite{AranPhamVenk2014}, rather than appearance features (such as SIFT \cite{Aran2008}, SURF \cite{BayEssTuytGool2008} or GLOH \cite{MikoSchm2004}). This is understandable considering that novelties of interest -- events and behaviours -- are inherently associated with motion, whereas appearance in a scene can vary greatly without the change being associated with anything of potential interest. For example, diurnal or seasonal changes in illumination can create major appearance differences while, on the other hand, resulting in no motion features due to their low temporal frequency; cars (or similarly pedestrians) passing down the road differ greatly one from another, yet it is reasonable to expect a much more constrained range of variability of their motion if they obey the rules of traffic.

In this paper we focus on the methods of the second group described above. We are particularly motivated to do so by their representation on the market on the one hand, and the lower amount of research attention they have attracted in comparison with the first group, on the other.

\vspace{-5pt}\subsection{Available data -- description and constraints}\vspace{-2pt}
The surveillance analysis methods we are interested in all start with the same procedure for feature extraction. As video data is acquired, firstly a dense optical flow field is computed. Then, to reduce the amount of data that needs to be processed, stored, or transmitted, a thresholding operation is performed. This results in a sparse optical flow field whereby only those flow vectors whose magnitude exceeds a certain value are retained; non-maximum suppression is applied here as well. Normal variability within a scene and subsequent novelty detection are achieved using this data.

One problem with the data pre-processing procedure described above is that it does not consider the effects of perspective distortion on the appearance of the scene. Most obviously, the thresholding applied on the optical flow vectors should be dependent on their location in the image plane. At present, the threshold is set sufficiently low to pick up potential motions of interest in the most distant parts of the scene, which has the disadvantage of unnecessarily increasing the amount of data extracted in the regions closer to the camera. Equally, any subsequent learning could be made more robust if it took the effects of perspective into account. Our goal in this paper is to empower the existing surveillance analytics with this information, that is, to estimate the dominant perspective scaling effect in a scene using the described sparse optical flow field.

In other words, our input data comprises a sequence of sets of motion vectors. Two temporally-neighbouring sets are separated by a uniform time interval $\Delta t$, which is governed by the frame rate of the camera (5--15~fps). Motion vector sets are in general of different sizes, and each set describes the sparse optical flow field at a particular time instant. All motion vectors are also spatially localized in the image plane.

\vspace{-5pt}\section{Proposed methods}\vspace{-2pt}
In the previous sections we explained why none of the existing methodologies for perspective estimation are adequate for the deployment in the operational setting of many CCTV systems. In this section we introduce two novel methods that solve the problem at hand effectively. The two solutions we propose can be contrasted by the spatial scale at which inference is made. The first algorithm adopts a dense approach, whereby a perspective estimate is made at all image plane locations and the final estimate for the whole scene emerges through a consensus of these estimates. The second algorithm operates at a larger scale. It divides the image plane into blocks and performs inference by accounting for motion statistics in different image blocks and the constraints between neighbouring blocks. At the bottom-most level, both methods are based on the observation that the motion of an object farther in a scene exhibits itself as a linearly scaled version of the apparent motion which would be observed if the object was closer to the camera. The key challenge arises from the fact that in practice it is impossible to perform such controlled calibration. Instead we formulate sets of constraints, of a different type for the dense algorithm and the block algorithm, which allow us to infer the perspective-induced scaling from the projected motion observed in the image plane.

\vspace{-5pt}\subsection{Dense approach -- pixel level}\label{ss:dense}\vspace{-2pt}
The first method we describe is dense in the sense that a local perspective estimation is made at every image locus (on the scale of a pixel). These local estimates are then polled and the overall estimate of the perspective distortion made through the consensus across the image plane. It is important to observe the assumption which is needed to make this approach sound. Specifically, we assume that in terms of its area when projected onto the image plane the scene is dominated by the ground plane in which motion takes place. This is a sensible assumption to make considering the purpose of CCTV cameras and the strategic manner of their placement.

We adopt the standard non-skew pin-hole camera model. Without loss of generality, if the origin of a right-hand coordinate system is placed at the focal point and the $x-y$ plane made parallel to the image plane at the focal length $f$, a 3D point $[x,y,z]^T$ is projected to the image plane point $[u,v]^T$ as follows:
{\small\begin{align}
  u=f \cdot k_u \cdot \frac{x}{z} && v=f \cdot k_v \cdot \frac{y}{z},
\end{align}}
where $k_u$ and $k_v$ are the camera's horizontal and vertical scaling parameters. The internal camera parameters -- namely its focal length $f$, and the scaling parameters $k_u$ and $k_v$ -- are of no interest to us since the normalization we seek is camera specific and is applied on a camera-by-camera basis.

Our goal is to estimate the dominant perspective-effected change in scale in the image plane. We formalize this by saying that we seek to estimate the quantity $\zeta$ we defined as:
{\small\begin{align}
  \zeta = \frac{dz/z}{dv}.
  \label{e:zeta}
\end{align}}
In other words, we wish to know how the distance of an object in the scene (or, equivalently, its perceived scale) changes with the change in its projected image locus. Here we show that $\zeta$ can be estimated from spatially and temporally local pure motion features (described in the previous section) only.

Consider the change in the vertical component of the projected (image plane) velocity of a point which corresponds to a world point moving in the ground plane. It is proportional to the component of the point's world velocity in the direction of the camera $\frac{dw}{dt}$ ($w$ is the projection of $z$ onto the ground plane) and inversely proportional to its distance $z$ from the camera's focal point. In other words:
{\small\begin{align}
  d\dot{v} = c \cdot \frac{dw}{dt} \cdot \left[ \frac{1}{z+dz}-\frac{1}{z} \right] = c \cdot \frac{dw}{dt} \cdot \frac{dz}{z^2},
  \label{e:ddotv}
\end{align}}
where we introduce the constant $c$ to `bundle' different constants for the sake of reducing clutter. By substitution in \eqref{e:zeta} we can derive:
{\small\begin{align}
  \zeta=\frac{dz/z}{dv}=\frac{z~d\dot{v}}{c \cdot \frac{dw}{dt} \cdot dv}.
  \label{e:zeta1}
\end{align}}
Similarly, since:
{\small\begin{align}
  \dot{v}=\frac{dv}{dt} = c \cdot \frac{dw/dt}{z},
\end{align}}
the expression in \eqref{e:zeta1} can be further simplified to:
{\small\begin{align}
  \zeta=\frac{dz/z}{dv}=\frac{d\dot{v}}{dv} \times \frac{1}{\dot{v}}
\end{align}}
Succinctly expressed, this result shows that in the context under consideration, perspective-caused scaling in the image plane can be estimated using the rate of change of velocity of a feature in uniform world motion across the image, normalized by its observed, image plane velocity.

\vspace{-5pt}\subsubsection{Model outliers}\vspace{-2pt}
Throughout our derivation in the preceding section it was assumed that the observed world motion was piecewise uniform. While it can be reasonably expected that this assumption will be satisfied in most cases, it is equally true that it will undoubtedly often be invalidated as well. For example, a walking pedestrian may increase the speed of walking upon spotting a bus, or stop because something has attracted his/her attention. However, these are not systemic outliers -- given sufficient data, it can be expected that opposite direction (acceleration vs.\ deceleration) deviations from the uniform motion will cancel each other out. A more serious challenge is posed by systemic outliers which emerge as a consequence of structural aspects inherent in the scene. For example, a corner in the road will consistently invalidate the assumption of uniform motion as cars entering the bend will slow down. Fortunately, this challenge is readily addressed in our framework considering that the dense nature of our method offers a large pool of quasi-independent perspective estimates. In particular, when polling different image locations, we aggregate all estimates in a vector, sort them and reject the top and bottom 15\% quantiles before computing the mean of the remainder to arrive at the global estimate.

\vspace{-5pt}\subsection{Coarse approach -- block level}\label{ss:coarse}\vspace{-2pt}
While successful in practice, as we will show in the next section, intuitively speaking the scale at which the method proposed in the previous section operates seems excessively fine considering the global nature of the assumption of one dominant perspective effect. This is further supported by the observation that in principle the same quantity $\zeta$ is estimated at every location in the image plane. Yet, in the proposed framework this polling is necessary as a means of rejecting outlier loci which do not satisfy the assumption of (on average) uniform world velocity of the corresponding features.

A simple way of reducing the number of image plane locations considered, and with it the computational cost of the method, could be implemented by prioritizing certain locations over others. For example, locations at which the amount of motion observed falls below a specified threshold could be reasonably considered as less reliable `voters' and thus rejected from being polled at all. However this approach still does not address the fundamental overkill which working on the pixel level appears to be.

\begin{figure}[thp]
\vspace{-5pt}
  \centering
  \subfigure[]{\includegraphics[width=0.21\textwidth]{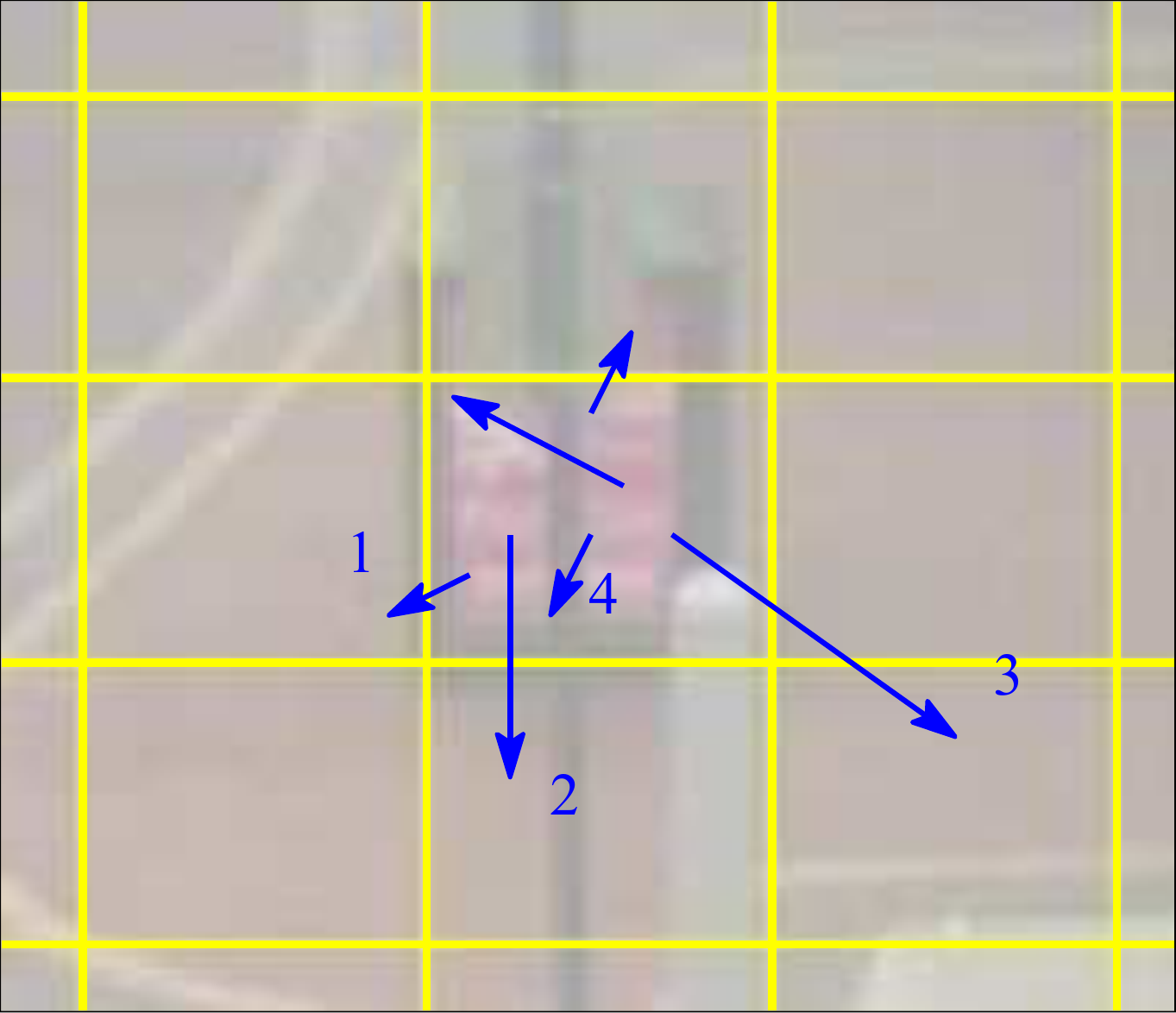}\label{f:coarse_a}}~~~~~~~~
  \subfigure[]{\includegraphics[width=0.21\textwidth]{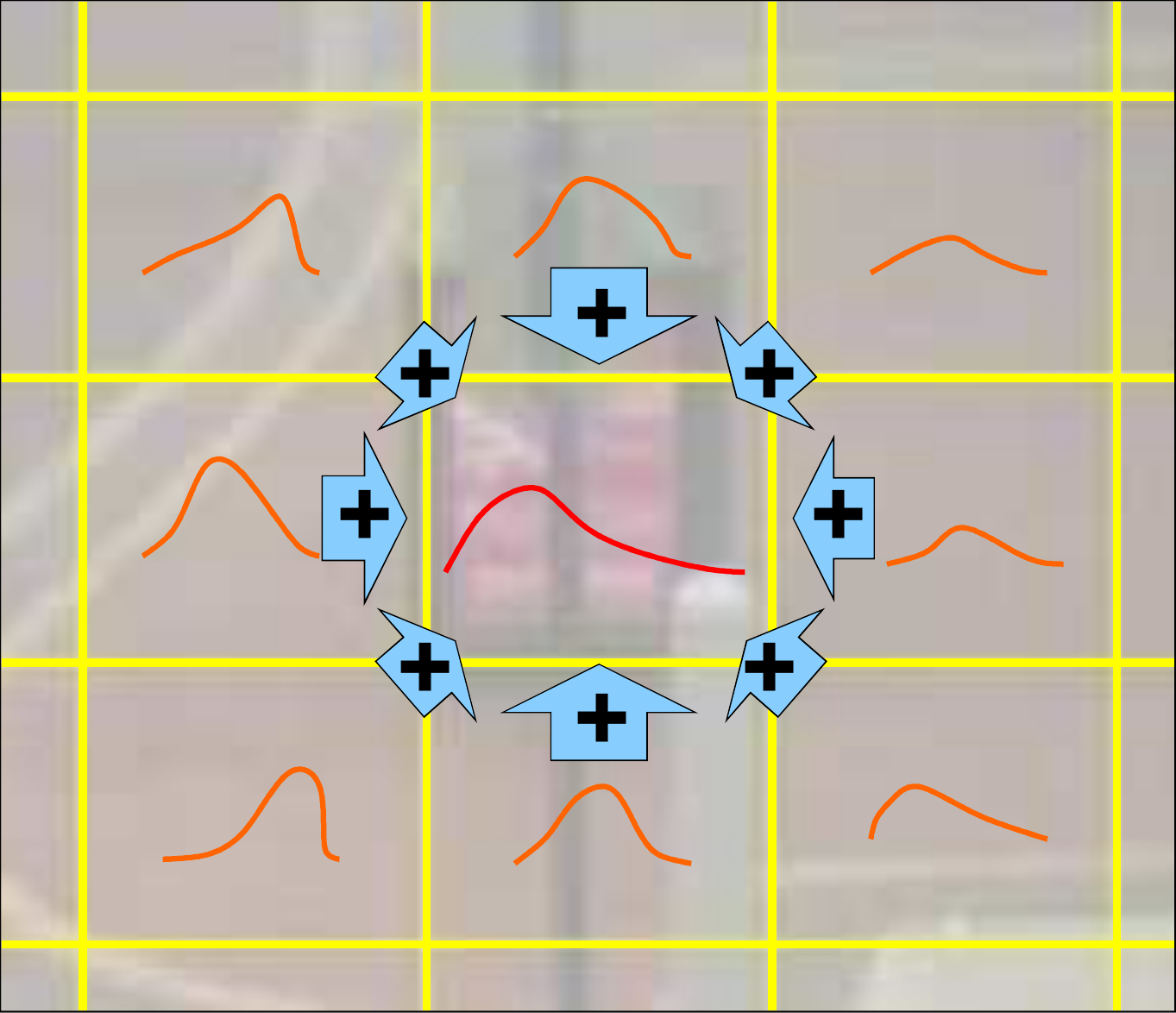}\label{f:coarse_b}}
  \vspace{-5pt}
  \caption{ Conceptual illustration of the proposed coarse algorithm. The key idea is to count motion vectors resulting in transitions between blocks into which the image is divided (a), and use thus estimate transition statistics together with the motion statistics within each block to formulate perspective scaling constraints between different blocks (b).}
\vspace{-5pt}\end{figure}

The method we describe next uses a division of the image plane into equally-sized rectangular blocks. This is illustrated in Figure~\ref{f:coarse_a}. Each block is treated as a unit, that is, any property pertaining to the block is associated with the block as a whole. The key idea underlying our approach is that motion statistics in different blocks can be related to one another. Most notably this is true for neighbouring blocks. We exploit this observation by recording nine measurements for each block. The first of these, $m_{i,j}$, is the average velocity magnitude within the block ($i$ and $j$ are the vertical and horizontal indices of the block). For example, in Figure~\ref{f:coarse_a} all of the shown motion vectors would contribute to the central block's mean motion magnitude; notice that we compute the mean of all magnitudes, rather than the magnitude of the mean motion vector. The remaining eight \emph{transition} measurements associated with the block, {\small$\rho^{(\ldots)}_{i,j}$}, quantify the relatedness of the motion within the block and the motions in the block's 8-neighbourhood. A specific {\small$\rho^{(\ldots)}_{i,j}$}, say {\small$\rho^{tl}_{i,j}$}, indicates the proportion of motion vectors which end in the block under the consideration and originate in its top-left neighbour (`tl'; similar indexing is used to denote the remaining directions, namely right and bottom, and the possible combinations). For example, in Figure~\ref{f:coarse_a} the motion vector labelled `1' contributes to {\small$\rho^{(r)}_{i,j-1}$}, the vector labelled `2' to {\small$\rho^{(t)}_{i+1,j}$}, the vector labelled `3' to {\small$\rho^{(l)}_{i,j+1}$} and {\small$\rho^{(tl)}_{i+1,j+1}$}, and the vector labelled `4' to no transition measurement at all (but it does contribute to $m_{i,j}$, of course).

Our idea is to formulate the problem of estimating the relative scaling between neighbouring rows of blocks in the form of an optimization task. Specifically, for the block indexed by $(i,j)$ we can write:
{\small\begin{align}
  m_{i,j}=&\rho_{i,j}^{(l)} ~m_{i,j-1} + \rho_{i,j}^{(r)}~m_{i,j+1} + \notag \\
          \Big[ &\rho_{i,j}^{(tl)}~m_{i-1,j} + \rho_{i,j}^{(t)}~m_{i-1,j} + \rho_{i,j}^{(tr)}~m_{i-1,j} \Big] \times \omega + \notag \\
          \Big[ &\rho_{i,j}^{(bl)}~m_{i+1,j} + \rho_{i,j}^{(b)}~m_{i+1,j} + \rho_{i,j}^{(br)} m_{i+1,j} \Big] \times \omega^{-1},
  \label{e:block}
\end{align}}
where $\omega$ is the factor quantifying perspective-effected scale change between consecutive rows of blocks. It is simple to show that it is related to the previously introduced $\zeta$ as:
{\small\begin{align}
  \zeta = \frac{1-\omega^{-1/2}}{h},
\end{align}}
where $h$ is the height of a block in pixels.

What \eqref{e:block} is saying is that the motion observed within $(i,j)$ is a mixture of motions in the neighbouring blocks, weighted by the proportional contribution of each block (via different {\small$\rho^{(\ldots)}_{i,j}$}) and where applicable the perspective-induced scaling $\omega$. This is illustrated conceptually in Figure~\ref{f:coarse_b}.

Since a constraint in the form of \eqref{e:block} can be written for each non-boundary block, thereby resulting in $(n-1) \times (m-1)$ equations with only a single unknown $\omega$, the system is over-determined. Thus we seek the solution which gives the least $L_2$-norm error for the error vector comprising the differences between the left and right-hand sides of \eqref{e:block} for different blocks. The optimal solution is readily obtained by employing one of a number of standard iterative schemes. Nevertheless, we found that a comparably accurate result could be obtained using an approximation which can be computed in the closed form. In particular we exploit the observation that the viewing setup of CCTV cameras is such that the scaling factor between neighbouring block rows $\omega$ will be close to 1. Thus writing $\omega$ as $\omega=1+\Delta \omega$ and using the first-order Taylor expansion whereby $\omega^{-1}$ can be approximated as $\omega^{-1} \approx 1 - \Delta \omega$ allows for \eqref{e:block} to be replaced by its approximate form:
{\small\begin{align}
  m_{i,j}\approx &\rho_{i,j}^{(l)} ~m_{i,j-1} + \rho_{i,j}^{(r)}~m_{i,j+1} + \rho_{i,j}^{(tl)}~m_{i-1,j} + \notag\\
                 &\rho_{i,j}^{(t)}~m_{i-1,j}  + \rho_{i,j}^{(tr)}~m_{i-1,j} + \rho_{i,j}^{(bl)}~m_{i+1,j} + \notag\\
                 &\rho_{i,j}^{(b)}~m_{i+1,j} + \rho_{i,j}^{(br)} m_{i+1,j}+\notag \\
                  \Big[ &\rho_{i,j}^{(tl)}~m_{i-1,j} + \rho_{i,j}^{(t)}~m_{i-1,j} + \rho_{i,j}^{(tr)}~m_{i-1,j} - \notag \\
                        &\rho_{i,j}^{(bl)}~m_{i+1,j} + \rho_{i,j}^{(b)}~m_{i+1,j} + \rho_{i,j}^{(br)} m_{i+1,j} \Big] \times \Delta \omega,
  \label{e:block2}
\end{align}}
Clearly this is now a set of linear equations which is readily solved in the closed form for $\Delta \omega$ which minimizes the $L_2$-norm error using the corresponding pseudo-inverse matrix.

\vspace{-5pt}\section{Evaluation and results}\label{s:eval}\vspace{-2pt}
To assess the effectiveness of the proposed algorithms, we evaluated their performance on nine large `real-world' data sets. It is important to emphasize that the data we used was not acquired for the purpose of the present work nor were the cameras installed with the same intention. Rather, we used data which was acquired using existing, operational surveillance systems. In particular, our data comes from five operational CCTV cameras in three major cities. Table~\ref{t:data} provides a summary the key statistics of the nine data sets, all of which were produced from original video feeds in $352 \times 288$ pixel resolution using the threshold of $1.5$ pixels to arrive at a sparse optic flow field from the initial dense computation using the algorithm of Lucas and Kanade~\cite{LucaKana1981}. For our coarse, block-based method we used a grid of $10\times 10$ blocks.

\begin{table}[htb]
\vspace{-10pt}
  \caption{Key statistics of the nine real-world data sets used in our evaluation. These were acquired from five operational CCTV cameras.
  }\vspace{5pt}
  \centering
  \small
  \renewcommand{\arraystretch}{1}
  \begin{tabular}{l|cccc}
  \Hline
  Scene       & Data    & Duration of      &  Frame & Avg.\ features \\
              & set     & data acquisition &  rate  & per frame\\
  \hline
  Scene 1 & 1       & 10~min &  5~fps  & 40.0\\
  Scene 2 & 1       & 2~h    & 15~fps  & 18.1\\
          & 2       & 2~h    & 15~fps  & 14.1\\
  Scene 3 & 1       & 2~h    & 15~fps  & 158.8\\
          & 2       & 2~h    & 15~fps  & 209.8\\
  Scene 4 & 1       & 2~h    & 15~fps  & 141.0\\
          & 2       & 2~h    & 15~fps  & 161.1\\
  Scene 5 & 1       & 2~h    & 15~fps  & 394.9\\
          & 2       & 2~h    & 15~fps  & 585.0\\
  \Hline
  \end{tabular}
  \label{t:data}
\end{table}

Considering that we could not obtain ground truth data directly, we used quasi-ground truth estimated using manually localized key points in the image plane. This was achieved by marking world-equidistant points at different distances from the camera. To increase the accuracy of thus obtained perspective estimates as well as to quantify the reliability of labelling, we asked for the labelling to be done by five different people for each scene.

\begin{figure}[!htb]
  \centering
  \subfigure[Dense algorithm]{\includegraphics[width=0.44\textwidth]{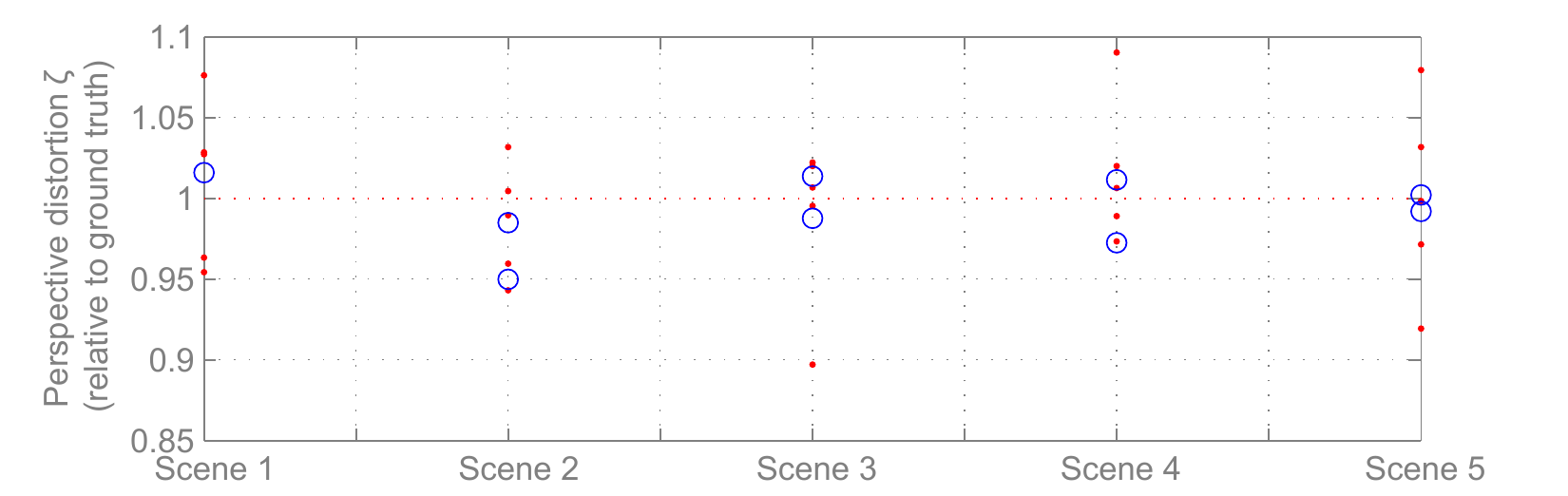}\label{f:resDense}}
  \subfigure[Coarse algorithm]{\includegraphics[width=0.44\textwidth]{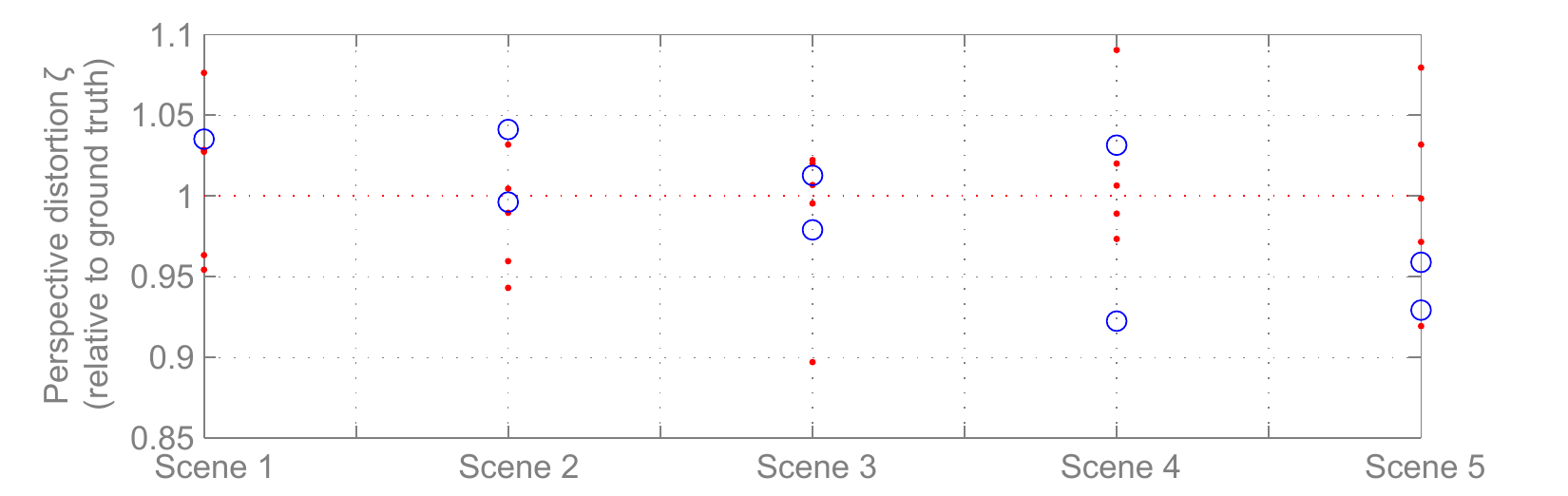}\label{f:resCoarse}}
  \caption{ Performance of the two proposed algorithms. Red dots are human estimates of $\zeta$; blue circles are  estimates by the proposed algorithm. Ordinate values are normalized so that the values of $\zeta$ for different scenes could be visualized on the same graph -- we applied scaling  which makes the mean value of the human estimate of $\zeta$ equal to unity.  }
  \vspace{-5pt}
\end{figure}

We started our evaluation by looking at the overall performance of the two proposed algorithms when applied to each of the nine data sets. A summary of the key results is shown in Figure~\ref{f:resDense} and Figure~\ref{f:resCoarse} for the dense and coarse model based algorithms respectively. Each small red dot corresponds to the estimate of the perspective distortion coefficient $\zeta$ estimated from the input of one human labeller of quasi-ground truth; thus there are five red dots for each scene, as we used five labellers. The blue circles represent the estimates produced by the proposed algorithm. There are two blue circles per scene for all scenes except for the first one (`Scene 1') as for all but this scene we had two data sets of motion vectors. Note that the ordinate values have been normalized so that the values of $\zeta$ for different scenes could be visualized on the same graph. Specifically, we applied scaling to the displayed values which makes the mean value of the estimate of $\zeta$ resulting from human labelling equal to unity. The first thing to observe from Figures~\ref{f:resDense} and~\ref{f:resCoarse} is the outstanding performance of both of our algorithms. This is witnessed by the proximity of all automatically produced estimates of $\zeta$ to unity which is well within the range of deviation of human-based estimates. Secondly, the accuracy of our algorithms, as well as the soundness of the premises underlying their derivation is further corroborated by their mutual agreement. Again, the difference in their output is smaller than the difference of estimates given by different humans. Quantitative analysis suggests that the performance of the dense approach somewhat exceeds that of the coarse alternative (by less than 5\%). While this is perhaps to be expected, considering the finer grained nature of the method and the robustness achieved by polling a large number of quasi-independent estimates, the difference is not significant.

\begin{figure*}[!htb]
  \centering
  \subfigure[Scene 1]{\includegraphics[width=0.33\textwidth]{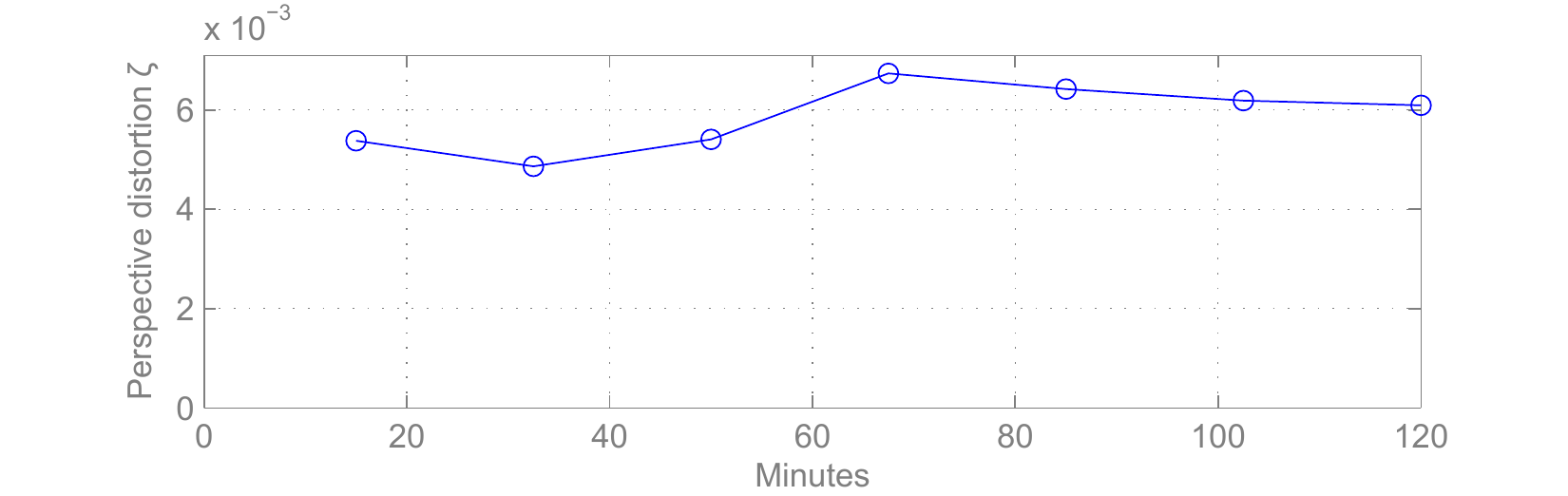}}
  \subfigure[Scene 2]{\includegraphics[width=0.33\textwidth]{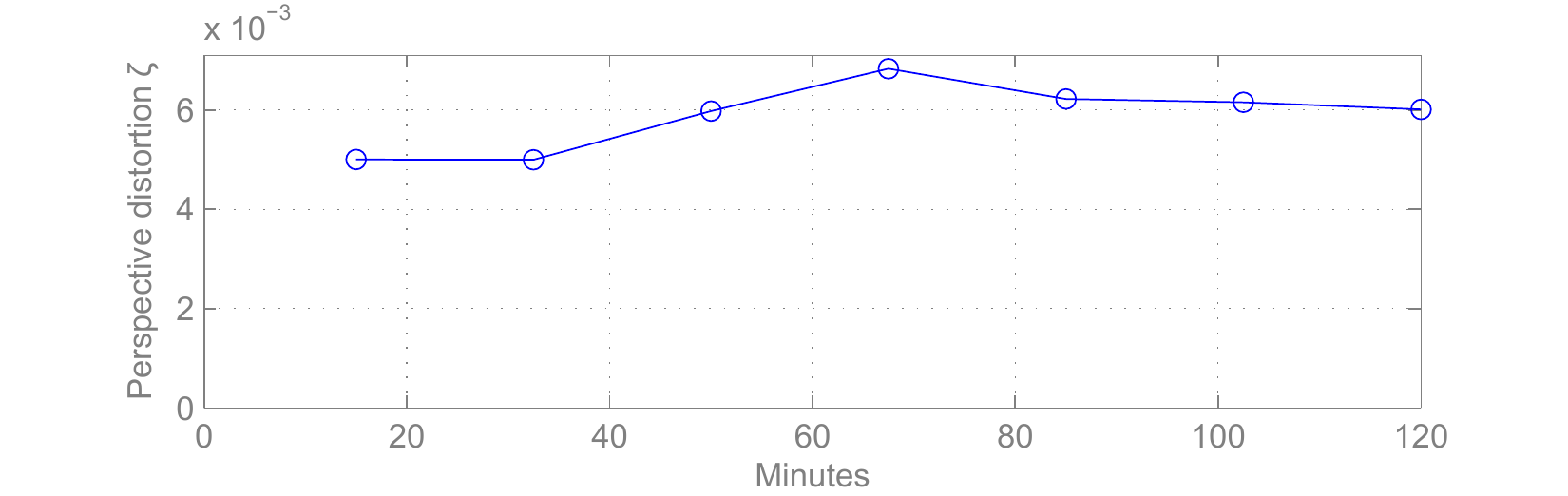}}
  \subfigure[Scene 3]{\includegraphics[width=0.33\textwidth]{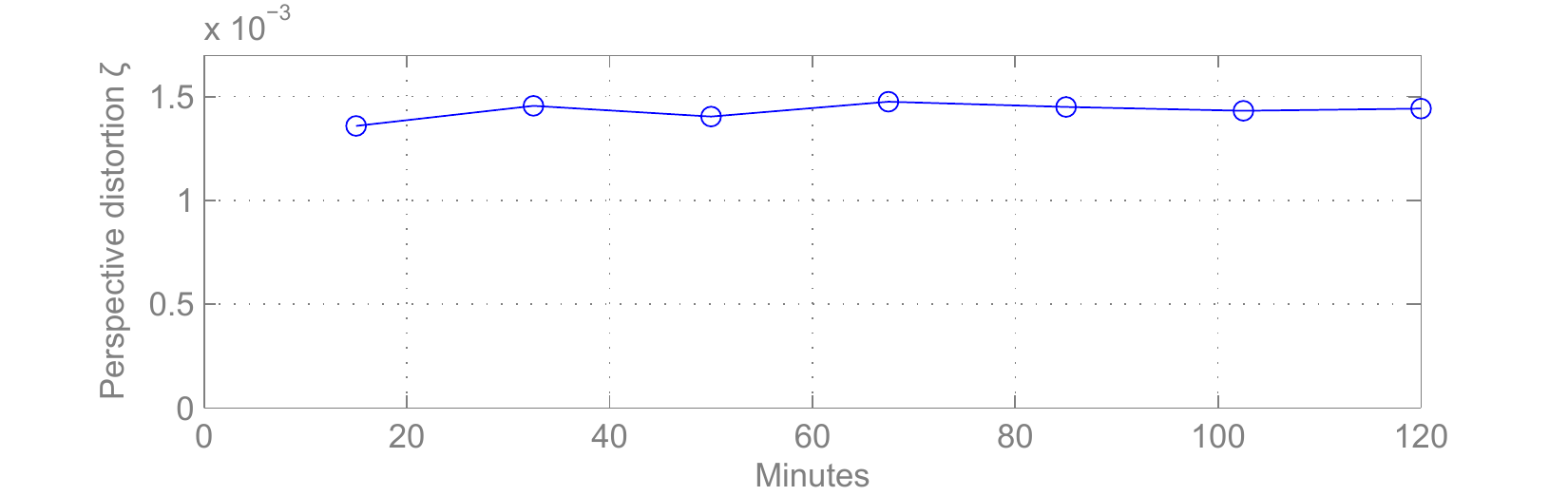}}
  \subfigure[Scene 4]{\includegraphics[width=0.33\textwidth]{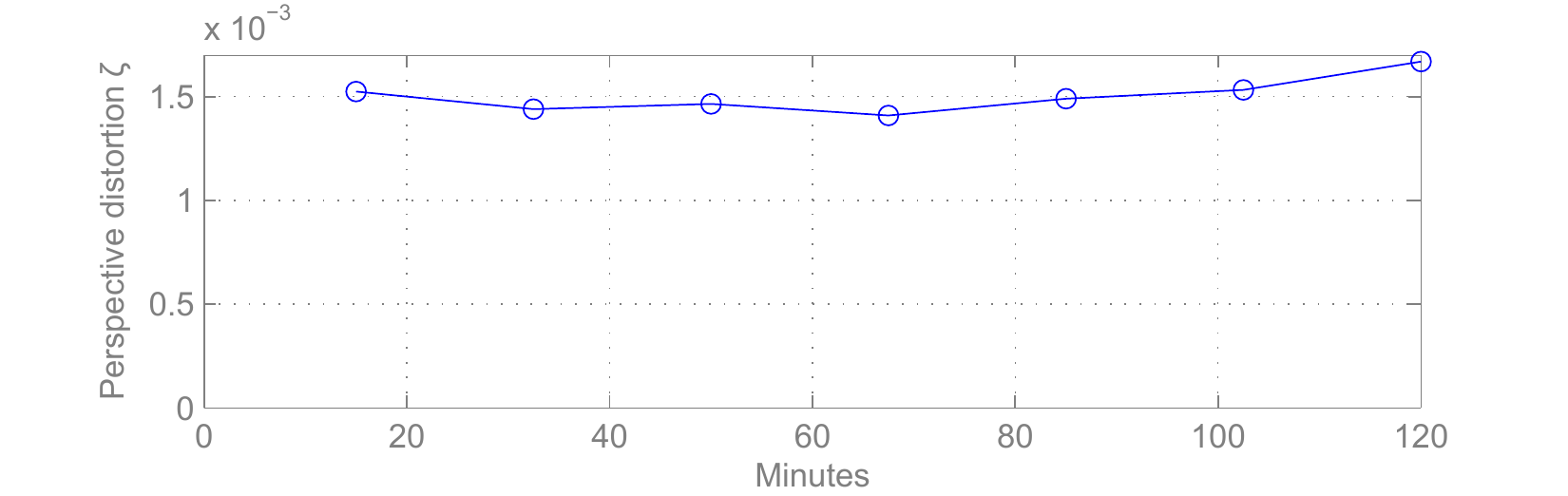}}
  \subfigure[Scene 5]{\includegraphics[width=0.33\textwidth]{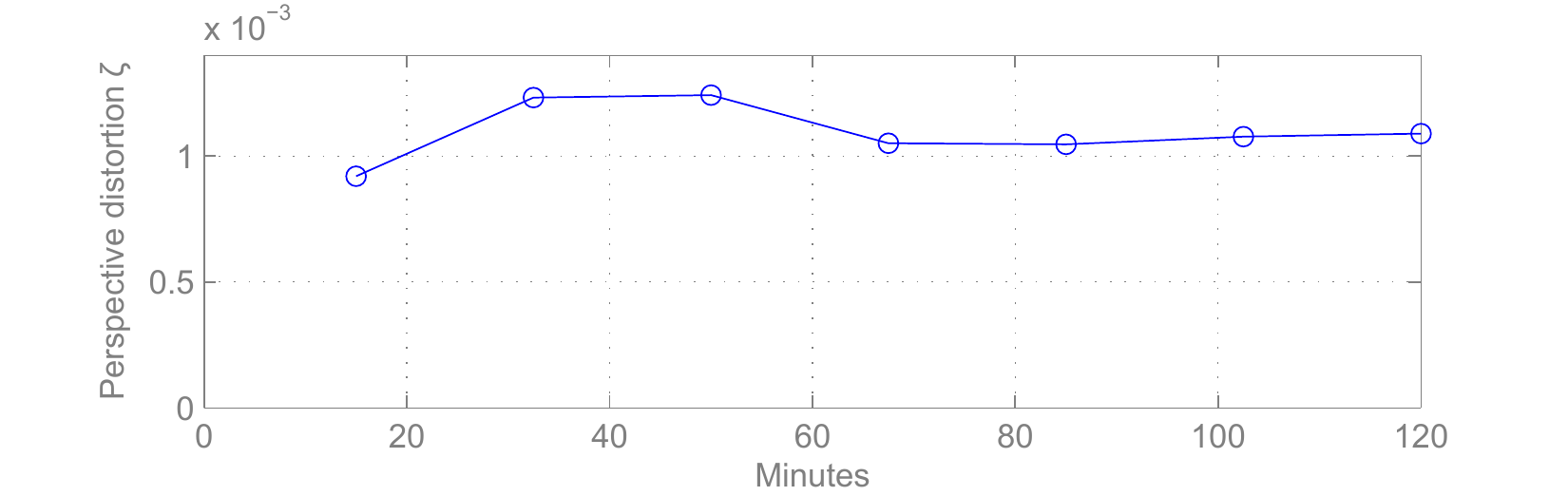}}
  \subfigure[Scene 6]{\includegraphics[width=0.33\textwidth]{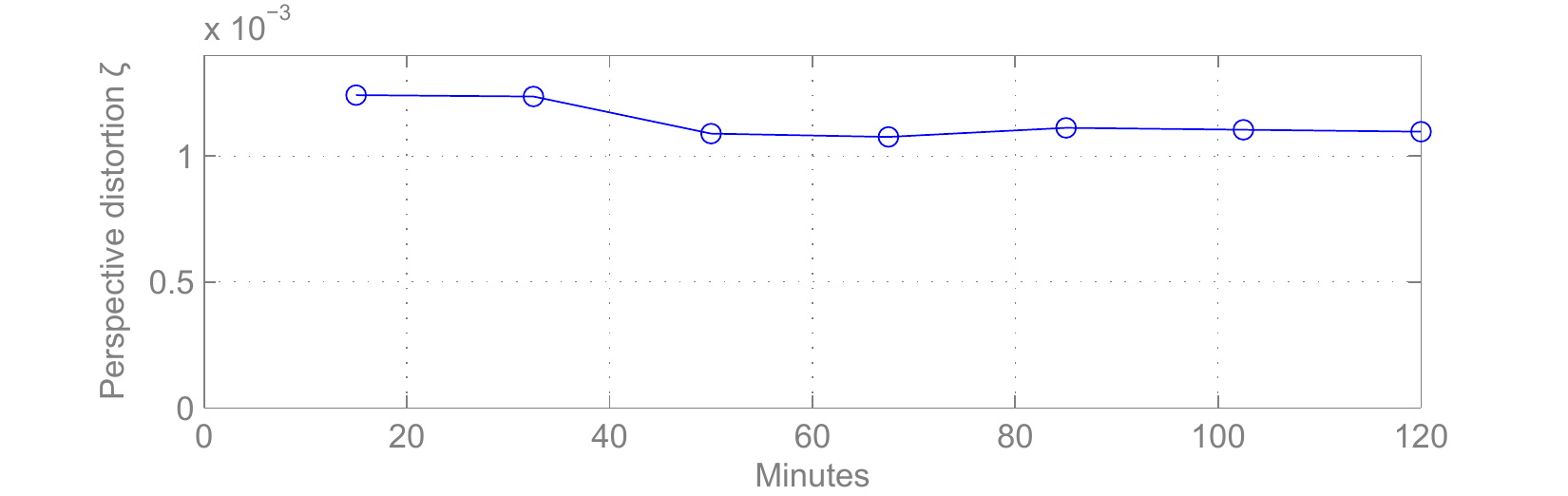}}
  \caption{ Variation of the scaling coefficient $\zeta$ with time i.e.\ with the accumulation of motion data. }
  \label{f:resTime}
\end{figure*}

In the next set of experiments we were interested in examining how much data is needed to reach a reliable estimate of $\zeta$. To do this, using each data set we produced seven different estimates -- the first one after using only the first 12.5\% of the data, the second one after using the first 25\% of the data, etc. The results are shown in Figure~\ref{f:resTime} (only the results obtained using the dense model-based algorithm are shown as the results of the coarse algorithm were so similar that they could not be displayed effectively on the same plots). As expected, the greatest inaccuracy, as well as the greatest change with newly acquired data, is exhibited at the beginning, when the least amount of data is available for inference. Remarkably, in all cases, the convergence towards the correct value is fast with the error lower than 5\% achieved within 90 minutes of data acquisition and in most cases in half of that time. For further analysis see~\cite{AranPhamVenk2015a}.

\small
\begin{spacing}{0.92}
  \bibliographystyle{IEEEbib}
  \bibliography{./my_bibliography1}

\begin{thebibliography}{10}

\bibitem{PhamAranVenk2015}
D.~Pham, O.~Arandjelovi{\'c}, and S.~Venkatesh,
\newblock ``Detection of dynamic background due to swaying movements from
  motion features.,''
\newblock {\em TIP}, 2015.

\bibitem{AranPhamVenk2015}
O.~Arandjelovi{\'c}, D.~Pham, and S.~Venkatesh,
\newblock ``Two maximum entropy based algorithms for running quantile
  estimation in non-stationary data streams.,''
\newblock {\em TCSVT}, 2015.

\bibitem{KahlHenr2007}
F.~Kahl and D.~Henrion,
\newblock ``Globally optimal estimates for geometric reconstruction
  problems.,''
\newblock {\em IJCV}, 2007.

\bibitem{HescRoum2011}
J.~A. Hesch and S.~I. Roumeliotis,
\newblock ``A direct least-squares ({DLS}) method for {PnP}.,''
\newblock {\em ICCV}, 2011.

\bibitem{ArmaArajSalv2003}
X.~Armangu, H.~Arajo, and J.~Salvi,
\newblock ``A review on egomotion by means of differential epipolar geometry
  applied to the movement of a mobile robot.,''
\newblock {\em PR}, 2003.

\bibitem{DomkAloi2007}
J.~Domke and Y.~Aloimonos,
\newblock ``A probabilistic framework for correspondence and egomotion.,''
\newblock {\em ICCVW}, 2005.

\bibitem{ChamCiptTanPham+2010}
T.-J. Cham, A.~Ciptadi, W.-C. Tan, M.-T. Pham, and L.-T. Chia,
\newblock ``Estimating camera pose from a single urban ground-view
  omnidirectional image and a {2D} building outline map.,''
\newblock {\em CVPR}, 2010.

\bibitem{BaziSeoDemoVass+2012}
J.-C. Bazin, Y.~Seo, C.~Demonceaux, P.~Vasseur, K.~Ikeuchi, I.~Kweon, and
  M.~Pollefeys,
\newblock ``Globally optimal line clustering and vanishing point estimation in
  {M}anhattan world.,''
\newblock {\em CVPR}, 2012.

\bibitem{HeroSzenZachDubs+2013}
A.~Herout, I.~Szentandr\'{a}si, M.~Zachari\'{a}s, M.~Dubsk\'{a}, and
  R.~Rudolf~Kajan,
\newblock ``Five shades of grey for fast and reliable camera pose
  estimation.,''
\newblock {\em CVPR}, 2013.

\bibitem{HoieEfroHebe2005}
D.~Hoiem, A.~A. Efros, and M.~Hebert,
\newblock ``Geometric context from a single image.,''
\newblock {\em ICCV}, 2005.

\bibitem{CornLeibCornVanG2008}
N.~Cornelis, B.~Leibe, K.~Cornelis, and L.~Van~Gool,
\newblock ``{3D} urban scene modeling integrating recognition and
  reconstruction.,''
\newblock {\em IJCV}, 2008.

\bibitem{DalaTrig2005}
N.~Dalai and B.~Triggs,
\newblock ``Histograms of oriented gradients for human detection.,''
\newblock {\em CVPR}, 2005.

\bibitem{MartAran2010}
R.~Martin and O.~Arandjelovi{\'c},
\newblock ``Multiple-object tracking in cluttered and crowded public spaces.,''
\newblock {\em ISVC}, 2010.

\bibitem{Aran2008}
O.~Arandjelovi{\'c},
\newblock ``Crowd detection from still images.,''
\newblock {\em BMVC}, 2008.

\bibitem{TranSoro2008}
D.~Tran and A.~Sorokin,
\newblock ``Human activity recognition with metric learning.,''
\newblock {\em ECCV}, 2008.

\bibitem{Phil}
{Philips Electronics N.V.},
\newblock ``A surveillance system with suspicious behaviour detection.,''
\newblock {\em Patent {EP1459272A1}}, 2004.

\bibitem{LaveKhanThur2007}
G.~Lavee, L.~Khan, and B.~Thuraisingham,
\newblock ``A framework for a video analysis tool for suspicious event
  detection.,''
\newblock {\em MTA}, 2007.

\bibitem{Aran2011a}
O.~Arandjelovi{\'c},
\newblock ``Contextually learnt detection of unusual motion-based behaviour in
  crowded public spaces.,''
\newblock {\em ISCIS}, 2011.

\bibitem{Inte}
intellvisions,
\newblock ``i{Q-P}risons,''
\newblock {\em {\url{http://www.intellvisions.com/}}}, Accessed March 2015.

\bibitem{AranPhamVenk2015d}
O.~Arandjelovi{\'c}, D.~Pham, and S.~Venkatesh,
\newblock ``The adaptable buffer algorithm for high quantile estimation in
  non-stationary data streams.,''
\newblock {\em IJCNN}, 2015.

\bibitem{iCet}
{iCetana},
\newblock ``{iMotionFocus}.,''
\newblock {\em {\url{http://www.icetana.com/}}}, Accessed March 2015.

\bibitem{AranPhamVenk2014}
O.~Arandjelovi{\'c}, D.~Pham, and S.~Venkatesh,
\newblock ``Stream quantiles via maximal entropy histograms.,''
\newblock {\em ICONIP}, 2014.

\bibitem{BayEssTuytGool2008}
H.~Bay, A.~Ess, T.~Tuytelaars, and L.~V. Gool,
\newblock ``{SURF}: Speeded up robust features.,''
\newblock {\em CVIU}, 2008.

\bibitem{MikoSchm2004}
K.~Mikolajczyk and C.~Schmid,
\newblock ``A performance evaluation of local descriptors.,''
\newblock {\em TPAMI}, 2004.

\bibitem{LucaKana1981}
B.D. Lucas and T.~Kanade,
\newblock ``An iterative image registration technique with an application to
  stereo vision.,''
\newblock {\em IJCAI}, 1981.

\bibitem{AranPhamVenk2015a}
O.~Arandjelovi{\'c}, D.~Pham, and S.~Venkatesh,
\newblock ``{CCTV} scene perspective distortion estimation from low-level
  motion features.,''
\newblock {\em TCSVT}, 2015.

\end{thebibliography}
\end{spacing}
\end{document}